%% file: main.tex
\documentclass{article}

\usepackage[nonatbib, final]{neurips_2024}
\usepackage{hyperref}
\usepackage{url}
\usepackage{colortbl}
\usepackage[table]{xcolor}
\definecolor{lightgray}{gray}{0.9}
\usepackage{mathtools}
\usepackage{arydshln}
\usepackage{multirow}
\usepackage{booktabs}
\usepackage[american]{babel}
\usepackage{threeparttable}
\usepackage{wrapfig}
\usepackage{natbib}
\usepackage{algorithm}
\usepackage{enumitem}
\usepackage{listings}
\usepackage[misc]{ifsym}
\usepackage[]{mdframed}
\usepackage{authblk}

\newcommand{\eg}[0]{\emph{e.g.}}

\title{One QuantLLM for ALL: Fine-tuning Quantized LLMs Once for Efficient Deployments
}

\author[1,2]{Ke Yi}
\author[4]{Yuhui Xu}
\author[3]{Heng Chang}
\author[3]{Chen Tang}
\author[3]{Yuan Meng}
\author[1]{Tong Zhang}
\author[2]{Jia Li}
\affil[1]{South China University of Technology }
\affil[2]{The Hong Kong University of Science
and Technology}
\affil[3]{Tsinghua University}
\affil[4]{Salesforce AI Research \authorcr \texttt{cs\_kerry@mail.scut.edu.cn}}

\begin{document}

\maketitle

\input{0_abstract}
\input{1_intro}

\input{2_relatedwork}
\input{3_methods}

\input{4_experiment}
\input{6_conclusion}
\clearpage
\bibliographystyle{apalike}
\bibliography{reference_fromscholar}
\end{document}

%% file: 0_abstract.tex
\begin{abstract}
Large Language Models (LLMs) have advanced rapidly but face significant memory demands.  While quantization has shown promise for LLMs, current methods typically require lengthy training to alleviate the performance degradation from quantization loss. However, deploying LLMs across diverse scenarios with different resource constraints, e.g., servers and personal computers, requires repeated training per application, which amplifies the lengthy training problem. Given that, it is advantageous to train a once-for-all (OFA) supernet capable of yielding diverse optimal subnets for downstream applications through one-shot training. Nonetheless, the scale of current language models impedes efficiency and amplifies interference from weight sharing between subnets. We make an initial attempt to extend the once-for-all framework to large language models. Specifically, we decouple shared weights to eliminate the interference and incorporate Low-Rank adapters for training efficiency. Furthermore, we observe the imbalance allocation of training resources from the traditional uniform sampling. A non-parametric scheduler is introduced to adjust the sampling rate for each quantization configuration, achieving a more balanced allocation among subnets with varying demands. We validate the approach on LLaMA2 families, and downstream evaluation confirms our ability to maintain high performance while significantly reducing deployment time faced with multiple scenarios.
\end{abstract}

%% file: 1_intro.tex
\section{Introduction}
Large Language Models have shown surprising performance in the past years. However, they suffer from huge storage and computational costs;
for example, inference with a LLaMA \citep{touvron2023llama} model with 70B parameters needs at least 280 GB of GPU memory. 
To further boost the LLMs development for fitting diverse scenarios, recent studies have adopted quantization to compress the model size and reduce the computational costs.

\begin{figure}[t]
    \centering
    \includegraphics[width=\linewidth]{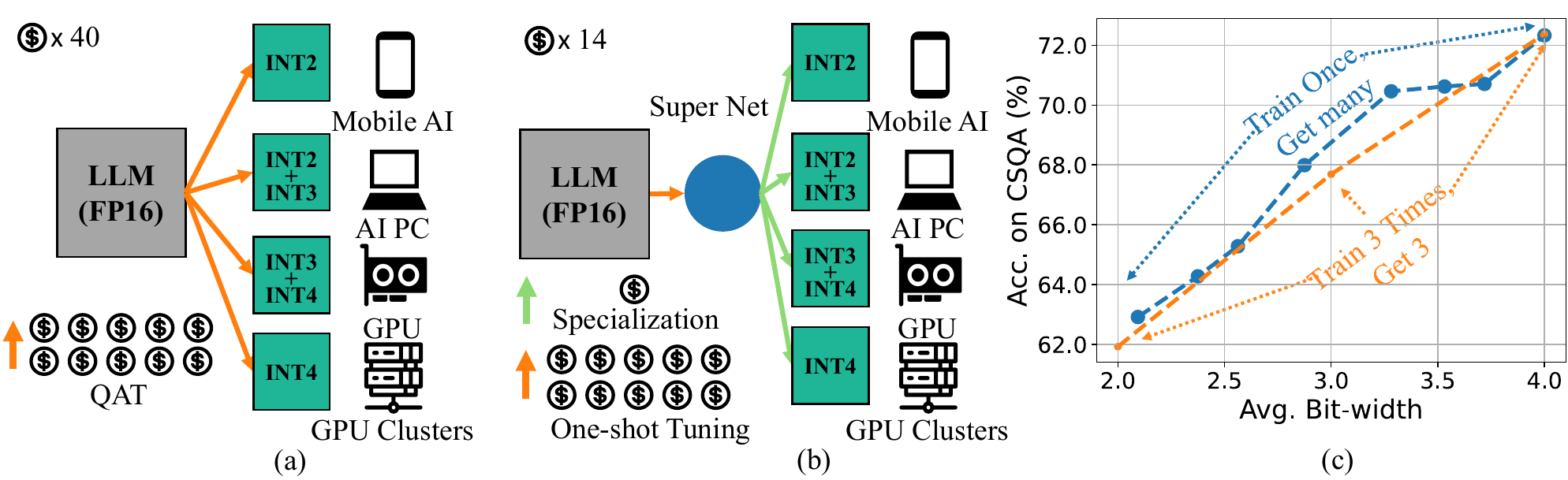}
    \caption{(a) Compressing Large Language Models (LLMs) for deployment across various platforms while ensuring performance is a challenging task. Applying Quantization-Aware Training (QAT) for each platform is both time-consuming and costly. (b) Instead, our objective is to one-shot fine-tune one quantized LLM that can be efficiently specialized for multiple platforms. The one-shot fine-tuning process significantly reduces the investment. (c) The LLM-QFA framework excels in swiftly delivering optimal networks under different resource constraints in one shot, whereas the traditional method requires repeated fine-tuning.}
    \label{fig: ofa}
\end{figure}

Previous works have extensively explored Post-Training Quantization \citep{frantar2022gptq,xiao2023smoothquant,lin2023awq} and Quantization-Aware Training \citep{dettmers2024qlora,xu2023qa} to alleviate the memory cost of LLMs. Post-training quantization (PTQ) offers swift model compression, albeit at the potential expense of performance. In contrast, Quantization-aware training (QAT) alleviates performance losses by simulating quantization errors during training, which is considerably more time-consuming than standard fine-tuning. 
When we need to deploy LLMs for diverse scenarios with different resource constraints, repeated quantization-aware training per scenario is unacceptable, as shown in Figure \ref{fig: ofa} (a). From the above analysis, the training major the cost of deployments; hence, it would be beneficial to train a once-for-all (OFA) supernet capable of delivering optimal subnets with diverse configurations (e.g., quantization bit-width) for each application, as shown in Figure \ref{fig: ofa} (b).

To the best of our knowledge, once-for-all quantization-aware training for LLMs has not been investigated, primarily due to the large scale of current language models and the high cost of traditional QAT. 
Previous researches based on once-for-all mainly utilize the weight-sharing strategy, which helps avoid model size explosion caused by allocating weight for each configuration \citep{wang2020hat,chen2021autoformer}. However, the weight-sharing combined with traditional QAT still has problems two-fold: 1) various quantization configurations (e.g., 2, 3, 4 bit-width) share the weight but have different orders of magnitude of quantization noise, resulting in the noteworthy interference problem and optimization challenges \citep{tang2024retraining}. 2) Tradition QAT is based on full-finetuning, combined with the time-consuming process of simulating quantization errors, which is inefficient even under the weight-sharing scheme.  

Furthermore, our observations reveal that the uniform sampling strategy used by traditional OFA brings an imbalance in the allocation of training resources. As illustrated in Figure \ref{fig: mixed gaussian}, subnets derived from uniform sampling exhibit a bias on their average bit-width, which falls into a low variance distribution. Consequently, subnets whose average bit-width deviates from this distribution are prone to under-fitting.

Integrating these aspects, we propose the \textbf{LLM-QFA} (Quantization-Aware Fine-tuning one LLM for All scenarios) framework that efficiently fine-tunes the once-for-all supernet for later yielding optimal subnets for diverse scenarios. First, we introduce interference-less fine-tuning to decouple the weights of different configurations, accompanied by Low-Rank adapters to enable efficient training. Specifically, we quantize the weights with different quantization configurations and freeze them, then we apply Low-Rank adapters to each quantized weight for later fine-tuning. Second, we propose a resource-balanced sampling strategy, which is based on a non-parametric scheduler that dynamically adjusts the sampling strategy across training steps.  

To evaluate our proposed framework, we conduct experiments on LLaMA2 models and validate the performance on the MMLU and Common Sense QA benchmarks. The results show that our proposed framework can yield diverse optimal quantized models for various scenarios. 
It is worth noting that our framework can be easily scaled up to even larger models since the training time per step is the same with previous LoRA-tuning \citep{xu2023qa}.
We summarize our contributions as follows:
\begin{itemize}[leftmargin=*]
\item We first introduce the once-for-all training paradigm for large language models (LLMs), which helps to reduce the training cost for deploying LLMs across diverse scenarios.
\item we decouple weights of configurations to mitigate interference issues and incorporate Low-Rank adapters to enhance the training efficiency.
\item To address the imbalance training caused by the uniform sampling strategy, we propose a resource-balanced sampling strategy that focuses on providing fair sampled opportunity across subnets with various resource demands.
\end{itemize}

%% file: 2_relatedwork.tex
\section{Related Work}
\paragraph{LLM Quantization}
Quantization is a compression technique that reduces the bit-width of weights and/or activations to save memory and accelerate inference. The quantization of LLM can be categorized into two main lines. The first one is post-training quantization (PTQ) \citep{frantar2022gptq, xiao2023smoothquant, lin2023awq,kim2023squeezellm}, which focuses on reducing the memory footprint without retraining. Although lots of designs are designed to mitigate the degradation of performance, \eg, handling outliers in parameters \citep{kim2023squeezellm,lillm} and dynamic quantization \citep{xiao2023smoothquant, lin2023awq}, PTQ still have to drop the ultra-low bit-width (\eg, 2 bit and 3 bit) to guarantee the performance. Hence, the second line, Quantization-Aware Training (QAT) can help alleviate the performance drop. The first QAT method applied on LLM \citep{liu2023llm} inherits the idea of traditional QAT, which is computationally expensive in the fine-tuning stage. To reduce the training cost, \citep{dettmers2024qlora, xu2023qa,guo2023lq,li2023loftq} utilizing LoRA-tuning on quantized weight and gain a decent performance. Specifically, \citep{xu2023qa} adds constraints on LoRA to maintain the quantization property after merging between LoRA weight and quantization weight, which firstly brings LoRA-tuning to actual quantization-aware training. Though Lora-tuning can save memory footprint and training costs, when faced with diverse development scenarios with different resource constraints, LoRA-tuning still falls into the pitfall of repeated training.

\paragraph{Once for All training}
Once-for-all training (OFA) methods \citep{wang2020hat,chen2021autoformer,yu2020bignas,tang2023elasticvit,tang2022arbitrary} aim to train a one-shot supernet that can serve diverse scenarios with different resource constraints and save expensive retraining per scenario. 
On non-LLMs, the success of one-shot training comes from the weight-sharing scheme between different configurations \citep{chen2021autoformer,yu2020bignas}, while weight-sharing also brings interference between different bit-widths for quantization-aware training \citep{tang2024retraining,tang2023elasticvit}. 
Moreover, traditional OFA with weight sharing necessitates fine-tuning entire parameters, which is impracticable for LLMs due to their extensive size.

%% file: 3_methods.tex
\section{Methodology}
\subsection{Problem definition}
\label{sec:Task-definition}
This paper focuses on the dimension of quantization to compress the LLMs for efficient deployment across diverse scenarios, which involves 1) post-training quantization to compress LLMs and 2) constructing the layer-wise mixed-precision supernet based on quantized LLMs and 3) optimizing the supernet.

\paragraph{Post-training Quantization} To reduce memory cost, it is effective to quantize the pre-trained weight of LLMs in low-bit representation; mathematically, given the bit-width $\mathbf{N}$ and the target weight $\mathbf{W}$, the quantization process can be defined as
\begin{small}
\begin{equation}
    \label{eq: round}
    \hat{\mathbf{W}} = \lfloor\frac{\mathbf{W}-\beta}{\alpha}\rceil, \alpha=(\max(\mathbf{W})-\min(\mathbf{W}))/(2^N-1) ,\beta=\min(\mathbf{W}),
\end{equation}
\end{small}

where $\alpha$ and $\beta$ are scaling and zero factors. $\lfloor\cdot\rceil$ denoted the rounding operation. $\hat{\mathbf{W}}$ is the quantized weight, and its elements are stored in a set of $\{0,1,\ldots,2^N-1\}$. 
Here, only two float point numbers and a series of integers are needed for storage and computation memory, 

\paragraph*{Layer-wise Mixed-precision Supernet}
In contrast to uniform bit-width quantization, mixed-precision quantization, which allows for varying bit-widths across different layers, can yield superior performance by capitalizing on the inherent redundancy in specific layers.
In this work, we build a supernet containing different quantization bit-width configurations layer-wisely. Each single path of the supernets denotes a mixed-precision LLM and we aim to optimize all single paths, which can be formulated as
\begin{equation}
    \label{eq: quantization}
    \{s_1,s_2,\dots,s_i,\dot,s_{N-1},s_{N}\}, \text{where}~ s_i = [Q_{1,i_1}, Q_{2,i_2}, \ldots, Q_{L,i_L}],
\end{equation}
where $s_i$ denotes one subnet. $L$ represents the number of layers in the large model. We quantize the model into $N$ different quantization bit-widths, denoted as $\textbf{B} = \{b_1, b_2, \dots, b_N\}$. $Q_{l,i}$ represent the quantized $l$-th layer with bit-width $b_i$. 
We apply quantize the pre-trained weight $\textbf{W}$ with {2, 3, 4} bit-width quantization. Hence, the quantity of subnets in the space is $3^{\textbf{L}}$. Our target is to 1) optimize all the subnets at once and 2) offer optimal subnets under given resource constraints.

\subsection{One-Shot Optimization}
\paragraph{Interference-Less Fine-tuning.}
We have observed that previous one-shot training methodologies \citep{cai2019once, yu2020bignas} gained success from their weight-sharing scheme, which avoids large model sizes caused by saving the weight of each configuration. However, the weight-sharing scheme also brings interference problems. Specifically, high and low bit-width have different quantization noise, and significantly superimposed quantization noise leads to optimization challenges \citep{tang2024retraining}. To alleviate interference between different configurations, the straightforward approach is to decouple shared weights and assign weights for each configuration, which is costly for large-size models.
\begin{figure}[t]
\centering
\includegraphics[width=\linewidth]{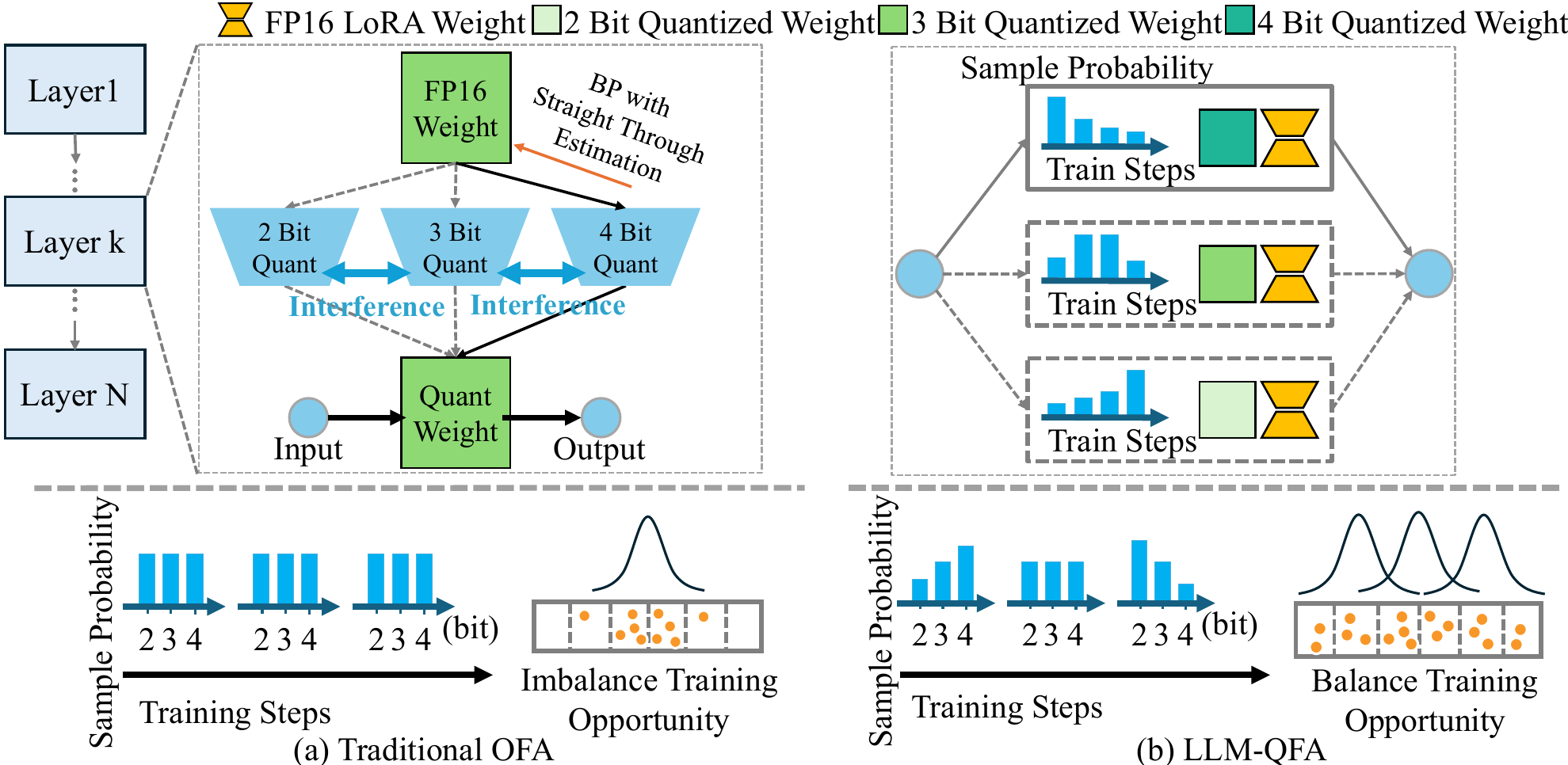}
\vspace{-3mm}
\caption{An illustration of the goal of LLM-QFA.
Compared with traditional OFA with Quantization-Aware Training, our approach circumvents interference issues by decoupling shared weight and incorporating the Low-Rank Adapter to further enhance the training efficiency. More notably, we employ a resource-balance sampling strategy to expedite the convergence of subnets across resource constraints.}
% An illustration of the goal of LLM-QFA. Compared with transitional OFA with Quantization-Aware Training, our approach avoids interference problems by decoupling sharing weight and integrating the Low-Rank Adapter to further enhance the training efficiency. More importantly, we take up resource-balance sampling strategy to accelerate the global convergence of subnets with different resource constraints.
\label{fig: main}
\end{figure}
Hence, we incorporate Low-Rank adapters to represent each quantization configuration, which only brings negligible extra cost compared with the size of LLMs. Specifically, the forward process can be defined as:
\begin{small}
\begin{equation}
    \label{eq: LoRA}
    \mathbf{Y}= {\alpha_i} \cdot \hat{\mathbf{W}_i} \cdot \mathbf{X} + {\beta_i} \cdot \mathbf{X} + \mathbf{B_i A_i} \cdot \mathbf{X},
\end{equation}
\end{small}
where ${\alpha_i},{\beta_i},\hat{\mathbf{W_i}}$ are factors and quantized weight under $i$-th bit-width configuration. ${\alpha} \cdot \hat{\mathbf{W}} + {\beta}$ is the dequantization process,  and $\mathbf{A}, \mathbf{B}$ denotes the weight of Low-Rank adapters. It is noticed that, during fine-tuning, only one of the Low-Rank adapters is updated, which is the key to avoiding interference between different configurations. 

To avoid heterogeneity between float point LoRA weights and quantized weight, which hinder the acceleration for inference, we follow QA-LoRA \citep{xu2023qa} to add constraints on adapters' weight for preserving quantization property after merging. 

Integrating the above designs, the task of optimizing all subnets can be formulated as 
\begin{small}
\begin{equation}
        \label{eq: task}
        \min_{\mathbf{W}_{L}} \sum_{a_i}  \mathcal{L}_{val}  \big( f(\mathbf{W}_{L},\mathbf{W}_{Q}, a_i) \big),
\end{equation}
\end{small}
where $f(\mathbf{W}_{L}, \mathbf{W}_{Q}, a_i)$ denotes the process that forms a sub-network according to architectural configuration $a_i$ and inherits corresponding quantization weight $W_{Q}$ and LoRA weight $W_{L}$. 

\paragraph{Resource-Balance Sampling Strategy.}

Fine-tuning all the subnets is a multi-objective problem. Given the impracticality of enumerating and tuning every subnet at each training iteration, a simplistic yet sub-optimal approach is to uniformly sample a few subnets from the configuration space for fine-tuning. 
Specifically, each layer has a uniform probability of choosing one quantization configuration, which can be formulated as $ \textbf{P}(Q_{l, i}) = \frac{1}{N}$.

\begin{figure}[t]
    \centering
    \includegraphics[width=\linewidth]{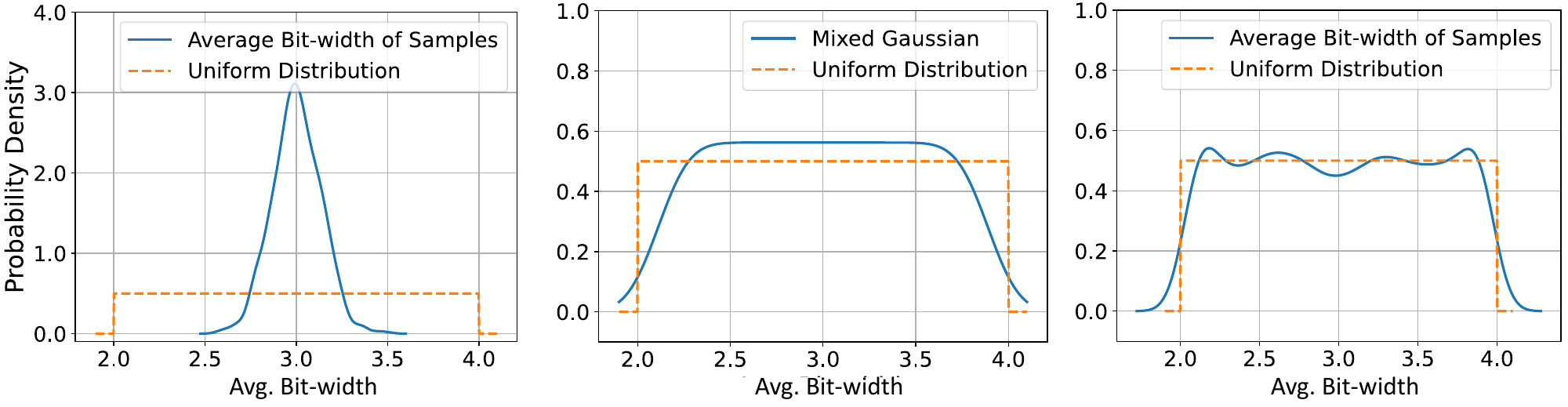}
    \vspace{-5mm}
    \caption{(a) Distribution of average bit-width of samples obtained from uniform sampling, approximating a low variance Gaussian distribution. (b) Mixed Gaussian Distribution can approximate Uniform Distribution. (c) Showcase of our Resource-Balance sampling strategy.}
    \label{fig: mixed gaussian}
\end{figure}

Though it seems fair, the naive uniform sampling strategy is biased toward subnets whose average bit-width is close to its expected value. Assume variable $q_i$ as quantization bit-width for $i_{th}$ layer. Variables [$q_1, q_2, \dots q_L$] are independent, hence the average of bit-width can be formulated as:
\begin{small}
\begin{equation}
    \begin{aligned}
        \label{eq: naive}
        \text{Var}[Bit(s)] & = \text{Var}[\frac{\sum_{i=1}^{L} q_i}{L}]
        = \frac{1}{L^2} \sum_{i=1}^{L} \text{Var}[q_i]
        = \frac{\sigma^2}{L},
\end{aligned}
\end{equation}
\end{small}
where the $Bit(s)$ denotes the average bit-width of the sampled and $\sigma^2$ denotes the variance of $q_i$. As shown in Figure \ref{fig: mixed gaussian} (a), the distribution of $Bit(s)$ is close to a normal distribution, where the variance is extremely small when $L=32$. Hence, the subnet with an average bit-width far from the distribution center would get unbalanced training resources. 
The negative impact of a uniform sampling strategy has not been studied previously. One of the reasons is that the weight-sharing scheme has all configurations updated frequently, though suffering from interference problems. Under the interference-less setting, weights are updated more sparsely; hence, the unbalanced training would lead to more pronounced under-fitting.

Revealed by Figure \ref{fig: mixed gaussian} (b), straightforwardly stacking normal distributions with different means can approximate a uniform distribution for $Bit(s)$ and alleviate the imbalance problem. From the implementation perspective, mixed Gaussian distribution can be achieved by setting different sampling strategies for configurations across training steps. The process can be formulated as 
\begin{equation}
    \begin{aligned}
        \label{eq: scheduler}
        \text{E}[Bit(s,t)] =  (b_N - b_1) \cdot \lvert 2 \cdot \frac{ t}{SL} - 1 \rvert ,
\end{aligned}
\end{equation}
where $SL$ is the length of one schedule epoch. $b_N$ represents the maximum bit-width, while in contrast, $b_1$ denotes the minimum bit-width. Within one schedule, the mean of distribution would move from $b_N$ to $b_1$ and then back to $b_N$, leading to a smooth switchover between schedule epochs. 

Compared to the uniform sampling strategy, our approach prevents bias on subnets in median size. Therefore, the subnet space converges more efficiently, which makes the following search process more effective. Compared to a shared-weight scheme, our approach can alleviate the interference problem with negligible extra memory costs. As a result, our approach provides a more efficient and effective way to optimize the Layer-wise Mixed-precision Supernet, which can be efficiently deployed in different scenarios with diverse resource constraints.

\subsection{Search Optimized Subnet}
We decouple the fine-tuning process and the searching process. No extra retraining cost is needed when finding the optimal subnet under the given resource constraint. The searching process starts with random searching, where a few subnets are sampled. Then, correlation analysis between the subnets' performance on the validation set and the quantization bit-width of each layer is conducted. Learning from the correlation, the sensitivity of each layer to quantization bit-width can be obtained and the search space can be further narrowed down. Finally, we further sample subnets from the narrowed search space, and the final optimal subnet is selected based on the performance of the validation set.

%% file: 4_experiment.tex
\section{Experiments}
\label{experiments}

\subsection{Settings}
\label{experiments:settings}

\paragraph*{Models and Quantization.} We conduct experiments on two LLMs, LLaMA2-7b and LLaMA2-13b. The quantization is based on GPTQ \citep{frantar2022gptq} with {2, 3, 4} bit-width quantization. The detailed quantization configuration, \eg, group size, and order, are consistent with QA-LoRA \citep{xu2023qa}.

\paragraph*{Datasets and Training Details.} We fine-tune models with Alpaca \citep{alpaca}, which contains 52K instruction-following data generated from GPT 3.5 \citep{wang2022self}. 
The length of one schedule epoch is 8k training steps.
Following previous works\citep{dettmers2024qlora, xu2023qa}, we use a paged AdamW optimizer with a batch size 16 and a learning rate of $2 \times 10^{-5}$. The training process is conducted on one A100 GPU, and only 8 GPU hours are needed to fine-tune one LLaMA2-7b-based supernet with 10K steps.

\paragraph{Evaluation.} We evaluate the performance of the models on MMLU \citep{hendrycks2021mmlu} and Common Sense QA benchmarks. The MMLU dataset contains four categories: Humanities, STEM, Social, and Other. The Common Sense QA benchmarks include HellaSwag \citep{zellers2019hellaswag}, PIQA \citep{bisk2020piqa}, WinoGrande \citep{sakaguchi2021winogrande}, ARC-e, ARC-c \citep{clark2018think}, BoolQ \citep{clark2019boolq}, and OBQA \citep{OpenBookQA2018}. 
For the MMLU Benchmark, we search the optimal subnets on the MMLU evaluation dataset. Initially, we sampled the first 100 subnets randomly and subsequently employed a shrinkage strategy to sample an additional 50 subnets, denoted as [100, 50]. For the Common Sense QA datasets, we similarly searched for optimal subnets on the ARC-C dataset with [100,50] setting. 
We report the $0$-shot and $5$-shot accuracy on MMLU and $5$-shot accuracy on Common Sense QA benchmarks.

\newcommand{\ours}{\textbf{LLM-QFA}}
\subsection{Main Results}
\begin{figure}[t]
    \centering
    \includegraphics[width=\linewidth]{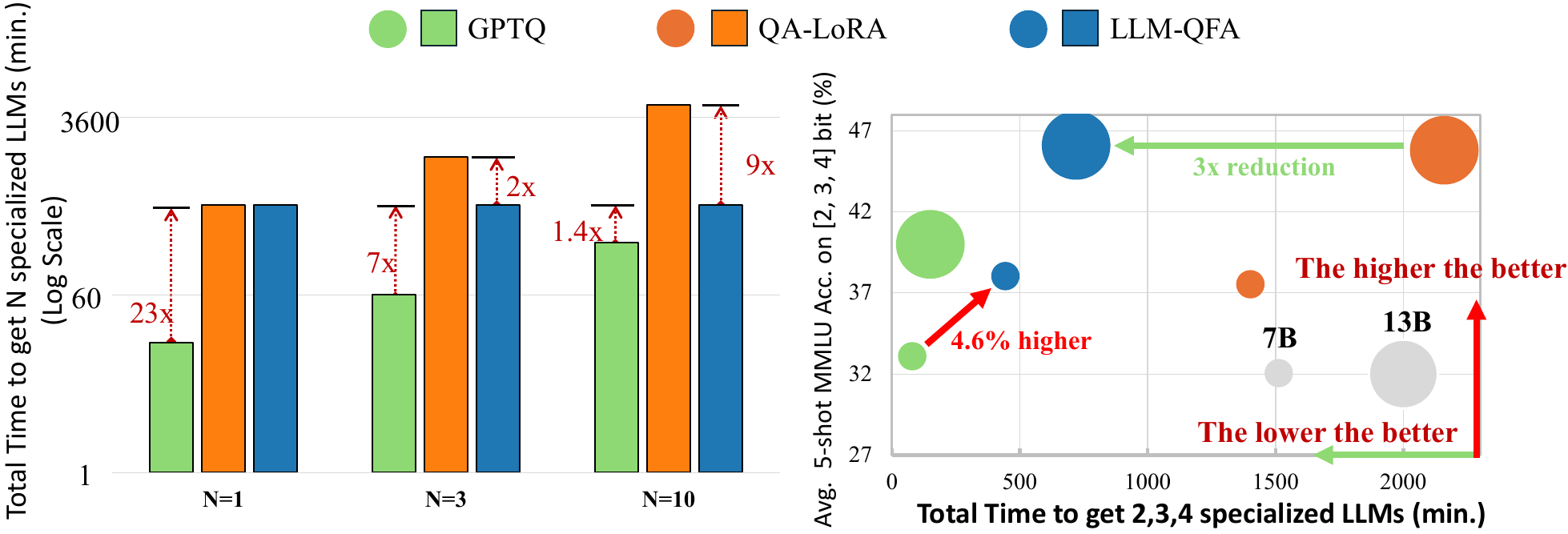}
    \vspace{-5mm}
    \caption{Left: The time required to obtain N specialized networks varies across methods. Our proposed QFA approach significantly reduces the time cost compared to the QA-LoRA method and achieves a comparable efficiency level to the pure quantization technique, GPTQ. Right: For each method, we obtain three specialized networks under (2, 3, 4) bit constraints on the LLaMA2-7b and LLaMA2-13B models. The average accuracy on the $5$-shot MMLU benchmark for networks quantized at (2, 3, 4) bits is reported. Although GPTQ can achieve a lower time cost, it is accompanied by an unacceptable level of performance degradation. Full results are provided in Table~\ref{tab: mmlu}.
    }
    \vspace{-5mm}
    \label{fig: efficiency}
\end{figure}
\begin{table}[!t]
    \renewcommand\arraystretch{0.6}
    \centering
    \caption{0-shot and 5-shot accuracy (\%) on the Massive Multitask Language Understanding (MMLU) dataset. Each block is based on the same foundation model specified in the first row. For each method, we present the metrics achieved under the bit-width resource constraints of {2, 3, 4}, as well as the corresponding averages.}
        \setlength{\tabcolsep}{0.8mm}
        {\resizebox{0.98\textwidth}{!}{
        \begin{tabular}{lc:ccccc:ccccc}
        \vspace{-1em}
        & & & & & & & & & &\\
        & & & & & & & & & &\\
        & & & & & & & & & &\\
        \toprule
        \multirow{2}{*}{\textbf{Method}} & \textbf{Bit} & \multicolumn{5}{c}{\textbf{MMLU} ($0$-shot)} & \multicolumn{5}{c}{\textbf{MMLU} ($5$-shot)} \\
        & \textbf{Const.} & \textbf{Hums.} & \textbf{STEM} & \textbf{Social} & \textbf{Other} & \textbf{Avg.} &\textbf{Hums.} & \textbf{STEM} & \textbf{Social} & \textbf{Other} & \textbf{Avg.} \\
        \midrule
            LLaMA2-7b & 16              & 48.3 & 35.2 & 48.8 & 45.8 & 43.6 & 51.6 & 37.3 & 52.2 & 49.9 & 46.8 \\ 
            \noalign{\vspace{0.1em}}\hdashline[0.8pt/1pt]\noalign{\vspace{0.1em}}
            GPTQ & 4                    & 40.4 & 33.7 & 45.9 & 42.2 & 39.9 & 50.5 & 36.9 & 50.5 & 47.5 & 45.1 \\ 
            GPTQ & 3 & 28.8 & 25.8 & 25.6 & 28.0 & 27.0 & 31.6 & 28.2 & 25.6 & 32.9 & 30.7 \\
            GPTQ & 2 & 23.8 & 23.7 & 22.5 & 23.8 & 23.5 & 24.3 & 23.0 & 23.9 & 26.1 & 24.2 \\
            \rowcolor{lightgray} GPTQ  & Avg. & ~ & ~ & ~ & ~ & 30.1  & ~ & ~ & ~ & ~ & 33.3  \\ 
            \noalign{\vspace{0.1em}}\hdashline[0.8pt/1pt]\noalign{\vspace{0.1em}}
            QA-LoRA & 4 & 49.7 & 37.5 & 51.4 & 47.8 & 45.7  & 49.8  & 36.8  & 49.8  & 47.8  & 45.1  \\
            QA-LoRA & 3 & 43.3 & 33.7 & 44.8 & 42.9 & 40.5  & 40.2  & 34.8  & 44.1  & 40.8  & 39.5  \\
            QA-LoRA & 2 & 32.6 & 27.2 & 35.6 & 33.2 & 31.7  & 27.2  & 26.9  & 29.0  & 30.5  & 28.3  \\
            \rowcolor{lightgray} QA-LoRA  & Avg. & ~ & ~ & ~ & ~ & 39.3  & ~ & ~ & ~ & ~ & 37.6  \\ 
            \noalign{\vspace{0.1em}}\hdashline[0.8pt/1pt]\noalign{\vspace{0.1em}}
            \ours{}  & 4 & 50.3 & 37.4 & 49.8 & 46.8 & 45.2  & 48.4  & 35.6  & 48.1  & 46.9  & 44.0  \\ 
            \ours{}  & 3 & 42.3 & 34.4 & 48.1 & 42.9 & 41.2  & 41.4  & 33.3  & 46.2  & 41.2  & 39.8  \\
            \ours{}  & 2 & 33.7 & 28.7 & 36.3 & 32.9 & 32.5  & 28.8  & 28.2  & 32.5  & 30.5  & 29.8  \\ 
             \rowcolor{lightgray} \ours{} & Avg. & ~ & ~ & ~ & ~ & \textbf{39.6}  & ~ & ~ & ~ & ~ & \textbf{37.9}  \\
            \midrule
            LLaMA2-13b & 16 & 56.9 & 42.4 & 61.0 & 55.6 & 52.8 & 62.9 & 44.4 & 63.9 & 56.7 & 55.7 \\ 
            \noalign{\vspace{0.1em}}\hdashline[0.8pt/1pt]\noalign{\vspace{0.1em}}
            GPTQ & 4 & 55.3 & 41.6 & 58.1 & 53.3 & 51.1 & 61.3 & 43.3 & 62.5 & 57.2 & 54.9 \\ 
            GPTQ & 3 & 42.0 & 31.8 & 43.6 & 41.3 & 39.0 & 41.4 & 36.5 & 46.7 & 43.7 & 41.5 \\ 
            GPTQ & 2 & 25.0 & 22.4 & 22.3 & 24.4 & 23.5 & 23.8 & 23.4 & 22.6 & 24.9 & 23.7 \\ 
            \rowcolor{lightgray} GPTQ & Avg. & ~ & ~ & ~ & ~ & 37.9  & ~ & ~ & ~ & ~ & 40.0  \\ 
            \noalign{\vspace{0.1em}}\hdashline[0.8pt/1pt]\noalign{\vspace{0.1em}}
            QA-LoRA & 4 & 56.9 & 41.5 & 60.4 & 54.9 & 52.3 & 59.6 & 42.7 & 62.2 & 57.4 & 54.2 \\ 
            QA-LoRA & 3 & 54.0 & 40.0 & 57.1 & 52.5 & 49.9 & 56.8 & 41.9 & 59.0 & 53.5 & 51.7 \\ 
            QA-LoRA & 2 & 32.6 & 28.9 & 31.4 & 35.3 & 31.8 & 30.3 & 28.2 & 34.4 & 36.5 & 32.0 \\ 
            \rowcolor{lightgray} QA-LoRA & Avg. & ~ & ~ & ~ & ~ & 45.3  & ~ & ~ & ~ & ~ & 45.8  \\ 
            \noalign{\vspace{0.1em}}\hdashline[0.8pt/1pt]\noalign{\vspace{0.1em}}
            \ours{} & 4 & 57.4 & 41.3 & 60.4 & 55.8 & 52.5 & 59.1 & 42.1 & 61.1 & 56.2 & 53.4 \\ 
            \ours{} & 3 & 56.3 & 40.3 & 58.8 & 54.6 & 51.3 & 56.7 & 40.6 & 59.9 & 54.5  & 51.8 \\ 
            \ours{} & 2 & 34.5 & 30.3 & 33.0 & 37.3 & 33.5 & 32.2 & 28.5 & 36.0 & 37.2 & 33.1 \\ 
            \rowcolor{lightgray} \ours{} & Avg. & ~ & ~ & ~ & ~ & \textbf{45.8}  & ~ & ~ & ~ & ~ & \textbf{46.1} \\ 
            \bottomrule
        \end{tabular}}}
        \label{tab: mmlu}
\end{table}
\paragraph{Comparisons with on MMLU.} 
Figure \ref{fig: efficiency} reports the comparison between LLM-QFA and Quantization-Aware training methods (QA-LoRA) and the Post-Training Quantization method (GPTQ) under (2, 3, 4) bit-widths. 
\ours{} demonstrates significantly higher efficiency than QA-LoRA faced with multiple deployment scenarios. This advantage stems from the training cost associated with LLM-QFA remaining constant, in contrast to the methods that scale linearly with the number of deployment scenarios \textbf{N}. 
Although our approach incurs a modestly higher time cost than GPTQ, the substantial performance degradation observed in GPTQ is unacceptable.
Table \ref{tab: mmlu} illustrates that, despite delivering only comparable performance under the 4-bit constraint, the average metrics of our method across (2, 3, 4) bit constraints consistently surpass those of QA-LoRA and GPTQ, without the need for costly repeated training.
\begin{figure}[t]
    \centering
    \includegraphics[width=0.8\linewidth]{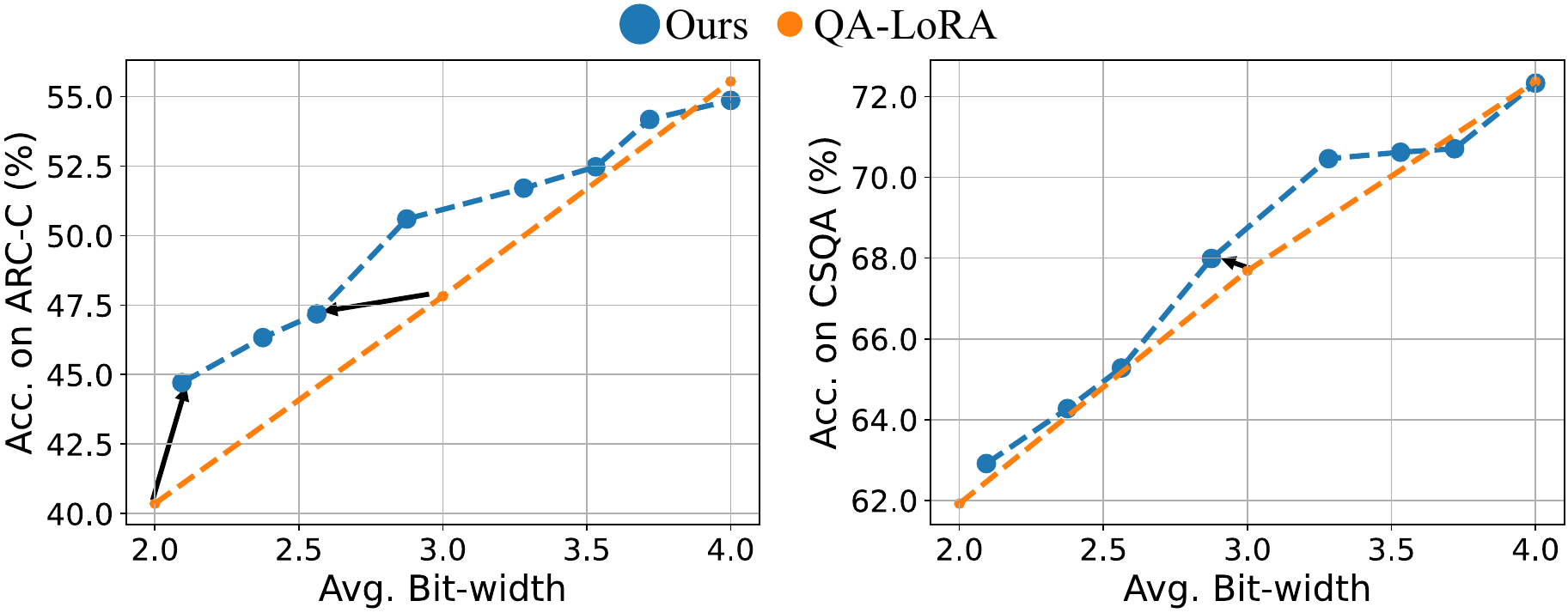}
    \vspace{-3mm}
    \caption{\ours{} can deliver multiple optimal subnets under different constraints. Left: Comparison of ARC-C dataset; Right: Comparison of the rest of Common Sense QA tasks.}
    \vspace{-6mm}
    \label{fig: qalora}
\end{figure}
\paragraph*{Comparisons on Common Sense QA.}
\begin{table}[!t]
    \renewcommand\arraystretch{1.0}
    \centering
    \caption{$5$-shot accuracy (\%) on the Common Sense QA tasks. Each block is based on the same foundation model specified in the first row. We organize all results under different quantization bit widths. Mixed precision configurations are searched on ARC-C, and the best configurations are tested on the rest of the Common Sense QA tasks.}
    \label{tab:csqa}
    \setlength{\tabcolsep}{1.2mm}
    \vspace{-5mm}
    {\resizebox{0.9\textwidth}{!}{
    \begin{tabular}{lc:c:ccccccccc}
        & & & & & & & & & &\\
        & & & & & & & & & &\\
    \toprule
    \multirow{2}{*}{\textbf{Method}} & \textbf{Bit} & \textbf{Eval} & \multicolumn{7}{c}{\textbf{Test}} \\
      & \textbf{Const.} & \textbf{ARC-C} & \textbf{HellaSwag} & \textbf{PIQA}  & \textbf{WinoGrande} & \textbf{ARC-e}  & \textbf{BoolQ} & \textbf{OBQA} & \textbf{Avg.} & \textbf{Std. (\%)} \\
    \midrule
    LLaMA2-7B    & 16 & 52.0 & 78.2 & 80.1 & 74.1 & 81.1 & 79.3 & 45.2 & 73.0 & 1.59\\
    \noalign{\vspace{0.1em}}\hdashline[0.8pt/1pt]\noalign{\vspace{0.1em}}   
    GPTQ  & 4               & 50.8 & 77.0 & 79.5 & 73.8 & 80.2 & 74.1 & 43.4 & 71.3 & 1.61 \\
    QA-LoRA     & 4         & 55.5 & 79.0 & 80.0 & 73.3 & 79.6 & 75.9 & 46.4 & \textbf{72.4} & 1.40 \\
    \ours{}    & 4          & 53.8 & 76.8 & 79.3 & 73.5 & 78.1 & 77.4 & 49.0 & 72.4  & 1.12\\
    \noalign{\vspace{0.1em}}\hdashline[0.8pt/1pt]\noalign{\vspace{0.1em}}   
    GPTQ  & 3                   & 30.1 & 49.9 & 68.3 & 59.3 & 55.5 & 44.3 & 35.0 & 52.1 & 1.13\\
    QA-LoRA     & 3             & 47.8 & 72.4 & 75.0 & 68.4 & 73.6 & 72.0 & 44.8 & 67.7 & 1.08 \\
    \ours{}     & 3             & 49.1 & 72.3 & 76.7 & 69.0 & 73.8 & 72.8 & 43.4 & \textbf{68.0} & 1.26\\
    \noalign{\vspace{0.1em}}\hdashline[0.8pt/1pt]\noalign{\vspace{0.1em}}  
    GPTQ    & 2                 & 25.8 & 26.2 & 51.1 & 50.6 & 26.0 & 41.7 & 25.0 & 36.8 & 1.31\\
    QA-LoRA     & 2             & 40.4 & 65.6 & 73.6 & 62.0 & 66.0 & 65.9 & 37.2 & \textbf{61.7} & 1.32\\
    \ours{}    & 2              & 43.1 & 64.8 & 73.2 & 62.2 & 67.0 & 64.3 & 38.8 & 61.7  & 1.16\\
    \midrule
    LLaMA2-13B    & 16          & 57.5 & 81.7 & 81.7 & 76.0 & 84.4 & 83.2 & 48.2 & 75.9 & 1.60\\
    \noalign{\vspace{0.1em}}\hdashline[0.8pt/1pt]\noalign{\vspace{0.1em}}  
    GPTQ  & 4                   & 56.5 & 81.1 & 80.9 & 75.6 & 83.3 & 81.7 & 47.4 & 75.0 & 1.58\\
    QA-LoRA     & 4             & 58.0 & 79.2 & 81.3 & 74.0 & 83.3 & 83.8 & 49.4 & 75.2 & 1.43\\
    \ours{}     & 4             & 56.0 & 79.6 & 82.0 & 73.2 & 83.5 & 83.2 & 51.0 & \textbf{75.4} & 1.31\\
    \noalign{\vspace{0.1em}}\hdashline[0.8pt/1pt]\noalign{\vspace{0.1em}}
    GPTQ  & 3                   & 47.8 & 68.6 & 77.7 & 67.9 & 77.1 & 71.9 & 42.8 & 67.7 & 1.38\\
    QA-LoRA & 3                 & 53.5 & 67.0 & 79.4 & 66.7 & 80.1 & 76.3 & 41.8 & 68.5 & 1.72\\
    \ours{}    & 3              & 53.7 & 75.1 & 79.7 & 70.3 & 80.5 & 78.4 & 48.0 & \textbf{72.0} & 1.27\\
    \noalign{\vspace{0.1em}}\hdashline[0.8pt/1pt]\noalign{\vspace{0.1em}}
    GPTQ  & 2                   & 27.8 & 25.8 & 50.2 & 50.2 & 26.6 & 37.8 & 23.4 & 35.7 & 1.26\\
    QA-LoRA     & 2             & 49.1 & 70.8 & 76.6 & 66.4 & 76.1 & 74.1 & 44.8 & 68.1 & 1.21\\
    \ours{}     & 2             & 49.2 & 70.9 & 77.0 & 67.2 & 76.3 & 74.3 & 44.6 & \textbf{68.4} & 1.24\\
    \bottomrule
    \end{tabular}}}
    \label{tab: csqa}
\vspace{-1em}
\end{table}

Table \ref{tab: csqa} reports the result of Common Sense QA. Consistent with the findings from the MMLU benchmark, 
LLM-QFA demonstrates comparable performance with baselines at extreme bit-width (2, 4) and outperforms at median bit-width (3). The advantage is significant with LLaMA2-13B under 3-bit constraints, where LLM-QFA gains 3.5\% accuracy improvement over QA-LoRA.

\paragraph*{\ours{} under Different Resource Constraints.}
Figure \ref{fig: qalora} summarizes the results of LLM-QFA under different bit-width constraints. LLM-QFA achieves 45.0\% ARC-C accuracy with 2.1 average bit-width, being 5\% more accurate than QA-LoRA with similar resource demands. Compared with QA-LoRA at 3-bit, our approach can achieve the same level of performance with fewer resources, a 1.2x reduction on ARC-C, and a 1.1x reduction on the rest of Common Sense QA.

\paragraph{Impact of Mixed Precision and Quality of Optimization.} 
\begin{wrapfigure}[18]{r}{0.45\textwidth}
\vspace{-0.5cm}
\centering
\includegraphics[width=0.45\textwidth]{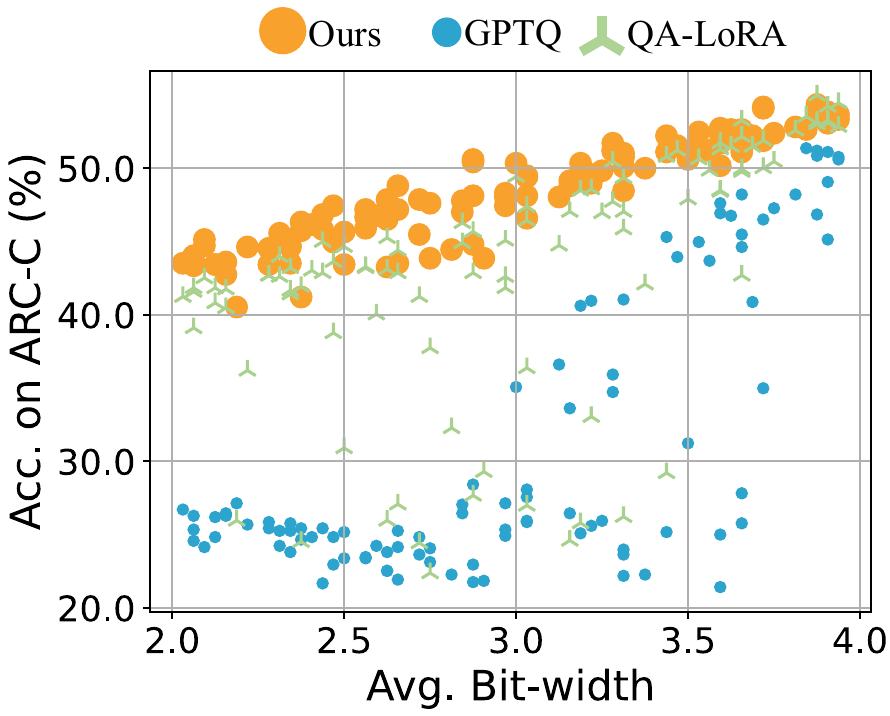}
\vspace{-2em}
\caption{Visualizing the degree of optimization by \ours. Subnets sampled from LLM-QFA show significant robustness over baselines with simple mixed-precision.}
\label{fig: mp}
\end{wrapfigure}
Previous results have significant performance improvement under the median resource constraints. To verify that the improvement does not only benefit from mixed precision, we separately sample 100 mixed-precision configurations for both GPTQ and QA-LoRA and evaluate them on the ARC-C dataset. To be noticed, we evaluate mixed-precision QA-LoRA based on the fine-tuned QA-LoRA weight at (2, 3, 4) bit. Figure \ref{fig: mp} demonstrates that our approach has a more robust performance across the dimension of resource demands, further validating that our method can help optimize all the subnets instead of only benefiting from the mixed-precision setting. Although the mixed-precision version of QA-LoRA exhibits a modest improvement in performance at higher bit-widths, it incurs a threefold increase in training time to achieve these results. Moreover, the observed performance instability suggests a potential loss of optimal subnet configurations under certain constraints.

\begin{figure}[t]
    \centering
    \vspace{-5mm}
    \includegraphics[width=0.8\linewidth]{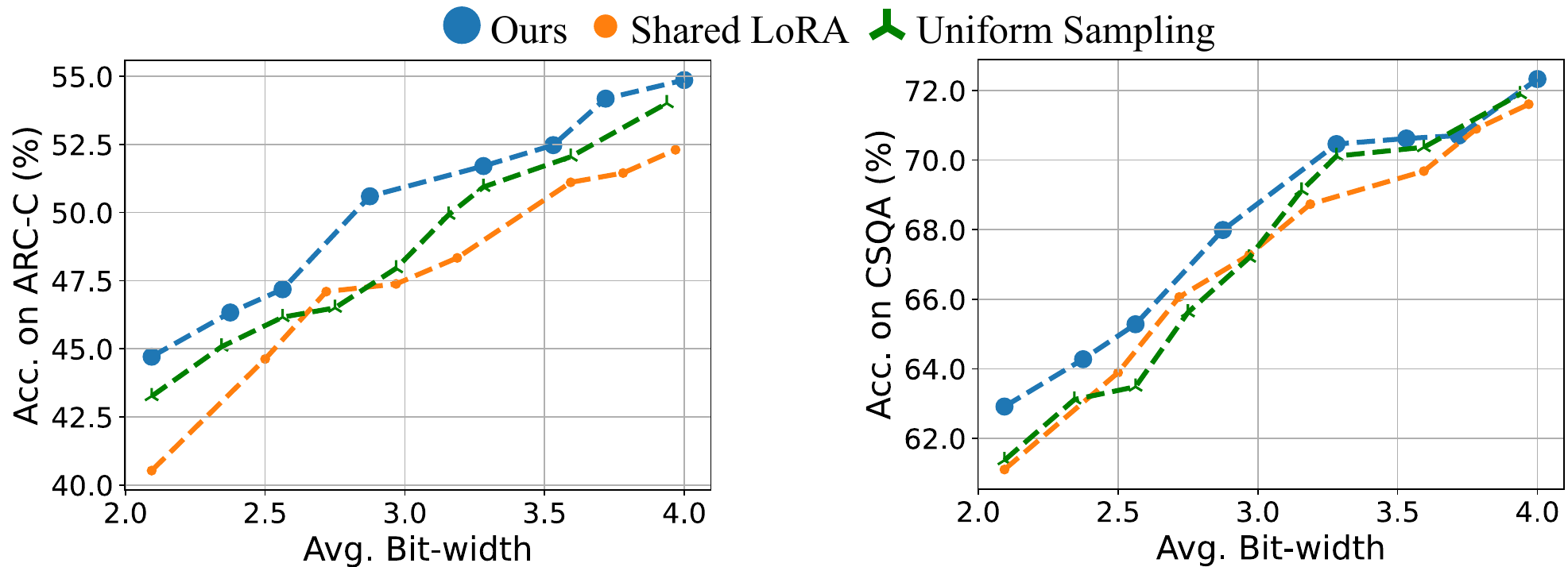}
    \caption{Verification of the effectiveness of Interference-Less Fine-Tuning and Resource-Balance Sampling Strategy.}
    \vspace{-4mm}
    \label{fig: uas}
\end{figure}
\begin{figure}[t]
    \centering
    \includegraphics[width=\linewidth]{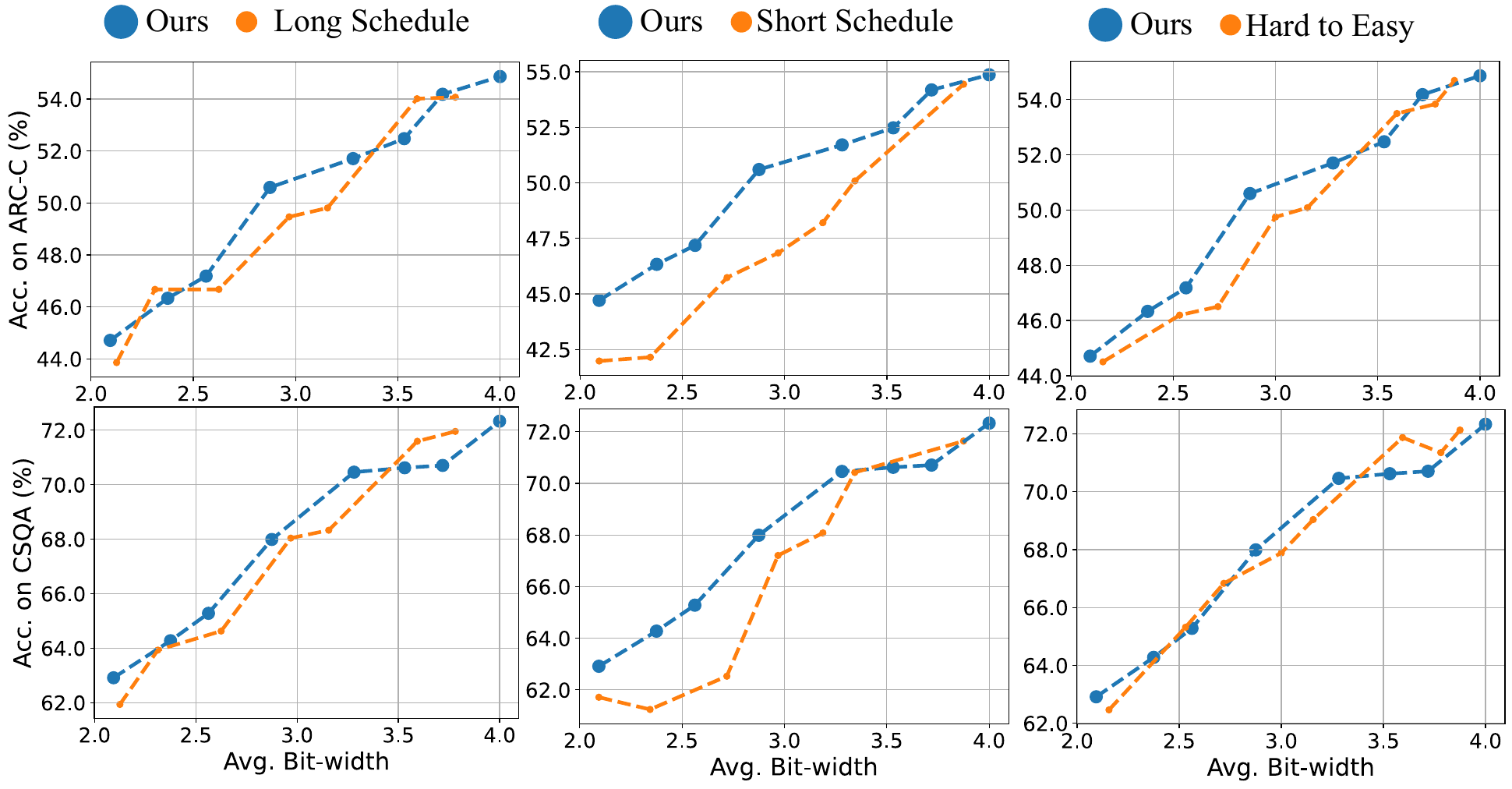}
    \vspace{-5mm}
    \caption{Common Sense QA accuracy (\%) of LLM-QFA with different scheduler settings.}
    \vspace{-5mm}
    \label{fig: ablation}
\end{figure}

\subsection{Ablation Study}
\paragraph{Ablation on Interference-Less Fine-tuning.}
To ascertain the effectiveness of decoupling shared weight, we introduce a variant of our method termed shared-LoRA,  wherein distinct quantization settings of a pre-trained weight share the same Low-Rank adapter. Figure \ref{fig: uas} reports that the shared-LoRA version fails the origin version across all resource demands, which validates the interference problem in one-shot training for LLMs. 
\paragraph{Ablation on Resource-Balance Sampling.}
Similarly, we implement a uniform sampling version of our method. Figure \ref{fig: uas} also shows a consistently under-performing uniform sampling strategy; even the resource-concentrated area (3 bit) falls short in the comparison. This has motivated the development of a resource-balanced sampling strategy for training, which is designed to counteract the challenges of under-fitting and over-fitting encountered in one-shot training.

\paragraph*{Ablation for Scheduler.}
Lastly, we investigate two aspects of configuration for the scheduler, which are the length of epochs (SL) and schedule orders.
In our main experiments, the epoch length is set to 8k training steps. For the short-term schedule, it is reduced to 1k steps, while for the long-term schedule, it is extended to 16k steps. Figure \ref{fig: ablation} demonstrates that the short-term diminishes robustness and hinders convergence, particularly at lower bit configurations.
Regarding the schedule orders, we initiate our training with 4-bit configurations, employing an easy-to-hard strategy. In this part, we assess the hard-to-easy setting. Figure \ref{fig: ablation} demonstrates that the order has negligible impact.

%% file: 6_conclusion.tex
\section{Conclusion}
This work introduces the \textbf{LLM-QFA} framework, a once-for-all Quantization-Aware training approach to reduce the training cost of deploying large language models (LLMs) across diverse scenarios. By decoupling the weights of different configurations and incorporating Low-Rank adapters, we enhance training efficiency and mitigate interference issues. A resource-balanced sampling strategy ensures fair training across subnets with various resource demands. Our experiments on LLaMA2 models show that \textbf{LLM-QFA} deliver optimal quantized models, demonstrating its effectiveness in reducing computational and storage costs while maintaining performance. Our framework can be easily scaled up to even larger models since the training time per step is the same as with previous LoRA tuning.

%% file: main.bbl
\begin{thebibliography}{}

\bibitem[Bisk et~al., 2020]{bisk2020piqa}
Bisk, Y., Zellers, R., Gao, J., Choi, Y., et~al. (2020).
\newblock Piqa: Reasoning about physical commonsense in natural language.
\newblock In {\em Proceedings of the AAAI conference on artificial intelligence}, volume~34, pages 7432--7439.

\bibitem[Cai et~al., 2019]{cai2019once}
Cai, H., Gan, C., Wang, T., Zhang, Z., and Han, S. (2019).
\newblock Once-for-all: Train one network and specialize it for efficient deployment.
\newblock {\em arXiv preprint arXiv:1908.09791}.

\bibitem[Chen et~al., 2021]{chen2021autoformer}
Chen, M., Peng, H., Fu, J., and Ling, H. (2021).
\newblock Autoformer: Searching transformers for visual recognition.
\newblock In {\em Proceedings of the IEEE/CVF international conference on computer vision}, pages 12270--12280.

\bibitem[Clark et~al., 2019]{clark2019boolq}
Clark, C., Lee, K., Chang, M.-W., Kwiatkowski, T., Collins, M., and Toutanova, K. (2019).
\newblock {BoolQ}: Exploring the surprising difficulty of natural yes/no questions.
\newblock {\em arXiv preprint arXiv:1905.10044}.

\bibitem[Clark et~al., 2018]{clark2018think}
Clark, P., Cowhey, I., Etzioni, O., Khot, T., Sabharwal, A., Schoenick, C., and Tafjord, O. (2018).
\newblock Think you have solved question answering? try arc, the ai2 reasoning challenge.
\newblock {\em arXiv preprint arXiv:1803.05457}.

\bibitem[Dettmers et~al., 2024]{dettmers2024qlora}
Dettmers, T., Pagnoni, A., Holtzman, A., and Zettlemoyer, L. (2024).
\newblock Qlora: Efficient finetuning of quantized llms.
\newblock {\em Advances in Neural Information Processing Systems}, 36.

\bibitem[Frantar et~al., 2022]{frantar2022gptq}
Frantar, E., Ashkboos, S., Hoefler, T., and Alistarh, D. (2022).
\newblock Gptq: Accurate post-training quantization for generative pre-trained transformers.
\newblock {\em arXiv preprint arXiv:2210.17323}.

\bibitem[Guo et~al., 2023]{guo2023lq}
Guo, H., Greengard, P., Xing, E.~P., and Kim, Y. (2023).
\newblock Lq-lora: Low-rank plus quantized matrix decomposition for efficient language model finetuning.
\newblock {\em arXiv preprint arXiv:2311.12023}.

\bibitem[Hendrycks et~al., 2021]{hendrycks2021mmlu}
Hendrycks, D., Burns, C., Basart, S., Zou, A., Mazeika, M., Song, D., and Steinhardt, J. (2021).
\newblock Measuring massive multitask language understanding.
\newblock In {\em International Conference on Learning Representations}.

\bibitem[Kim et~al., 2023]{kim2023squeezellm}
Kim, S., Hooper, C., Gholami, A., Dong, Z., Li, X., Shen, S., Mahoney, M.~W., and Keutzer, K. (2023).
\newblock Squeezellm: Dense-and-sparse quantization.
\newblock {\em arXiv preprint arXiv:2306.07629}.

\bibitem[Li et~al., 2023a]{lillm}
Li, S., Ning, X., Hong, K., Liu, T., Wang, L., Li, X., Zhong, K., Dai, G., Yang, H., and Wang, Y. (2023a).
\newblock Llm-mq: Mixed-precision quantization for efficient llm deployment.
\newblock {\em Efficient Natural Language and Speech Processing Workshop at Advances in Neural Information Processing Systems}.

\bibitem[Li et~al., 2023b]{li2023loftq}
Li, Y., Yu, Y., Liang, C., He, P., Karampatziakis, N., Chen, W., and Zhao, T. (2023b).
\newblock Loftq: Lora-fine-tuning-aware quantization for large language models.
\newblock {\em arXiv preprint arXiv:2310.08659}.

\bibitem[Lin et~al., 2023]{lin2023awq}
Lin, J., Tang, J., Tang, H., Yang, S., Dang, X., and Han, S. (2023).
\newblock Awq: Activation-aware weight quantization for llm compression and acceleration.
\newblock {\em arXiv preprint arXiv:2306.00978}.

\bibitem[Liu et~al., 2023]{liu2023llm}
Liu, Z., Oguz, B., Zhao, C., Chang, E., Stock, P., Mehdad, Y., Shi, Y., Krishnamoorthi, R., and Chandra, V. (2023).
\newblock Llm-qat: Data-free quantization aware training for large language models.
\newblock {\em arXiv preprint arXiv:2305.17888}.

\bibitem[Mihaylov et~al., 2018]{OpenBookQA2018}
Mihaylov, T., Clark, P., Khot, T., and Sabharwal, A. (2018).
\newblock Can a suit of armor conduct electricity? a new dataset for open book question answering.
\newblock In {\em EMNLP}.

\bibitem[Sakaguchi et~al., 2021]{sakaguchi2021winogrande}
Sakaguchi, K., Bras, R.~L., Bhagavatula, C., and Choi, Y. (2021).
\newblock Winogrande: An adversarial winograd schema challenge at scale.
\newblock {\em Communications of the ACM}, 64(9):99--106.

\bibitem[Tang et~al., 2024]{tang2024retraining}
Tang, C., Meng, Y., Jiang, J., Xie, S., Lu, R., Ma, X., Wang, Z., and Zhu, W. (2024).
\newblock Retraining-free model quantization via one-shot weight-coupling learning.
\newblock {\em arXiv preprint arXiv:2401.01543}.

\bibitem[Tang et~al., 2022]{tang2022arbitrary}
Tang, C., Zhai, H., Ouyang, K., Wang, Z., Zhu, Y., and Zhu, W. (2022).
\newblock Arbitrary bit-width network: A joint layer-wise quantization and adaptive inference approach.
\newblock In {\em Proceedings of the 30th ACM International Conference on Multimedia}, pages 2899--2908.

\bibitem[Tang et~al., 2023]{tang2023elasticvit}
Tang, C., Zhang, L.~L., Jiang, H., Xu, J., Cao, T., Zhang, Q., Yang, Y., Wang, Z., and Yang, M. (2023).
\newblock Elasticvit: Conflict-aware supernet training for deploying fast vision transformer on diverse mobile devices.
\newblock In {\em Proceedings of the IEEE/CVF International Conference on Computer Vision}, pages 5829--5840.

\bibitem[Taori et~al., 2023]{alpaca}
Taori, R., Gulrajani, I., Zhang, T., Dubois, Y., Li, X., Guestrin, C., Liang, P., and Hashimoto, T.~B. (2023).
\newblock Stanford alpaca: An instruction-following llama model.
\newblock \url{https://github.com/tatsu-lab/stanford_alpaca}.

\bibitem[Touvron et~al., 2023]{touvron2023llama}
Touvron, H., Lavril, T., Izacard, G., Martinet, X., Lachaux, M.-A., Lacroix, T., Rozi{\`e}re, B., Goyal, N., Hambro, E., Azhar, F., et~al. (2023).
\newblock Llama: Open and efficient foundation language models.
\newblock {\em arXiv preprint arXiv:2302.13971}.

\bibitem[Wang et~al., 2020]{wang2020hat}
Wang, H., Wu, Z., Liu, Z., Cai, H., Zhu, L., Gan, C., and Han, S. (2020).
\newblock Hat: Hardware-aware transformers for efficient natural language processing.
\newblock {\em arXiv preprint arXiv:2005.14187}.

\bibitem[Wang et~al., 2022]{wang2022self}
Wang, Y., Kordi, Y., Mishra, S., Liu, A., Smith, N.~A., Khashabi, D., and Hajishirzi, H. (2022).
\newblock Self-instruct: Aligning language model with self generated instructions.
\newblock {\em arXiv preprint arXiv:2212.10560}.

\bibitem[Xiao et~al., 2023]{xiao2023smoothquant}
Xiao, G., Lin, J., Seznec, M., Wu, H., Demouth, J., and Han, S. (2023).
\newblock Smoothquant: Accurate and efficient post-training quantization for large language models.
\newblock In {\em International Conference on Machine Learning}, pages 38087--38099. PMLR.

\bibitem[Xu et~al., 2023]{xu2023qa}
Xu, Y., Xie, L., Gu, X., Chen, X., Chang, H., Zhang, H., Chen, Z., Zhang, X., and Tian, Q. (2023).
\newblock Qa-lora: Quantization-aware low-rank adaptation of large language models.
\newblock {\em arXiv preprint arXiv:2309.14717}.

\bibitem[Yu et~al., 2020]{yu2020bignas}
Yu, J., Jin, P., Liu, H., Bender, G., Kindermans, P.-J., Tan, M., Huang, T., Song, X., Pang, R., and Le, Q. (2020).
\newblock Bignas: Scaling up neural architecture search with big single-stage models.
\newblock In {\em Computer Vision--ECCV 2020: 16th European Conference, Glasgow, UK, August 23--28, 2020, Proceedings, Part VII 16}, pages 702--717. Springer.

\bibitem[Zellers et~al., 2019]{zellers2019hellaswag}
Zellers, R., Holtzman, A., Bisk, Y., Farhadi, A., and Choi, Y. (2019).
\newblock Hellaswag: Can a machine really finish your sentence?
\newblock {\em arXiv preprint arXiv:1905.07830}.

\end{thebibliography}
